\documentclass{article} 
\usepackage{iclr2026_conference,times}

\usepackage{amsmath,amsfonts,bm}

\def\eqref#1{equation~\ref{#1}}

\def\1{\bm{1}}

\DeclareMathAlphabet{\mathsfit}{\encodingdefault}{\sfdefault}{m}{sl}
\SetMathAlphabet{\mathsfit}{bold}{\encodingdefault}{\sfdefault}{bx}{n}

\usepackage{hyperref}
\usepackage{url}
\usepackage{graphicx}%

\usepackage{tabularx}
\usepackage{array}
\usepackage{ragged2e}

\newcolumntype{Y}{>{\RaggedRight\arraybackslash}X}

\title{Physics-Constrained Neural Networks for Improved Short-Term Weather Forecasting: A Case Study over the South Pacific}

\author{Egor Bugaev, Fedor Buzaev, Dmitry Efremenko, Denis Derkach \& Fedor Ratnikov \\
Faculty of Computer Science\\
Higher School of Economics\\
Moscow, Russia \\
\texttt{\{ebugaev,fa.buzaev\}@hse.ru} \\
}

\iclrfinalcopy 
\begin{document}

\maketitle

\begin{abstract}
This study introduces enhancements to physics-constrained neural networks (PCNNs) that improve the accuracy and stability of hybrid short-term weather forecasting models. Building on the WeatherGFT architecture, three innovations are proposed. First, an upgraded numerical solver—combining a fifth-order weighted essentially non-oscillatory scheme (WENO-5), a $\beta$-plane approximation, and subgrid-scale viscosity—permits a fourfold increase in the integration time step (to 1200 s) while reducing the daily mean squared error by up to 26\%. Second, a unified autoregressive hybrid block replaces the original chain of 24 specialised modules, eliminating overfitting to specific lead times; and third, the physical core is integrated with two state-of-the-art neural backbones, resulting in PI-PredFormer and PI-IAM4VP. Evaluation on the WeatherBench South Pacific subset (2000–2004) shows that these hybrids reduce root mean squared error at 1–12 h lead times by 8–22\% compared to purely neural counterparts, while better preserving physical consistency. These results demonstrate that incremental refinement of hybrid components offers a practical route toward more accurate and efficient short-range weather forecasting.
\end{abstract}

\section{Introduction}
Accurate short-term (1–24 h) weather forecasts are critical for safeguarding lives and infrastructure in sectors including aviation, maritime transport, and emergency response~\citep{Molteni1996-ph,Vitart2021,Han2022}. While Numerical Weather Prediction (NWP) models generate physically consistent forecasts by integrating governing dynamical equations, they suffer from high computational costs and rely on empirical parameterizations for sub-grid-scale processes \citep{Zngl2023}. In contrast, pure machine learning (ML) models learn statistical patterns from vast datasets, offering forecasts orders of magnitude faster but often violating fundamental physical laws \citep{Grazzini2024, VahidYousefnia2024}.

A promising path forward lies in hybrid physics-constrained neural networks (PCNNs), which embed the residuals of governing PDEs directly into the learning objective. A seminal architecture in this domain is WeatherGFT \citep{WEATHER_GFT}, which interleaves Transformer blocks with a partial differential equation (PDE) solver. In parallel, highly efficient neural backbones have emerged for spatio-temporal prediction. Notably, PredFormer \citep{PredFormer} and IAM4VP \citep{imvp} achieve state-of-the-art accuracy with lightweight architectures, making them ideal for real-time applications. However, these models lack built-in physical guarantees, which can lead to unrealistic states.

In this study, we advance the hybrid modeling paradigm by combining the physical rigor of physics-constrained frameworks with the expressiveness of recent spatiotemporal architectures. Rather than proposing a wholly new architecture, we systematically enhance WeatherGFT and transfer its physical core to powerful, lightweight neural backbones. Our specific contributions are:

\begin{enumerate}
    \item \textbf{Enhanced solver.}  
    Central-difference schemes are replaced by a fifth-order weighted essentially non-oscillatory scheme \mbox{WENO-5} \citep{Stevens2020} with a $\beta$-plane Coriolis term and sub-grid viscosity, increasing the integration step from 300\,s to 1200\,s while reducing 24-h mean squared error by up to 26\,\%.

    \item \textbf{Universal hybrid block.}  
    The WeatherGFT original chain of 24 specialized modules is collapsed into a \emph{single, hourly PCNN block} with a learnable weight~$\gamma$ that adaptively balances data-driven and physics-based updates, eliminating over-fitting to pre-set lead times.

    \item \textbf{Integration pathways.} Two new hybrids are developed:
      \begin{itemize}
        \item Physics-informed PredFormer (PI-PredFormer), which applies post-hoc physics correction to the temporal Transformer PredFormer outputs every 15 min;
        \item Physics-informed IAM4VP (PI-IAM4VP), which incorporates the PCNN block into IAM4VP with physics corrections after each hourly forecast.
      \end{itemize}
\end{enumerate}

Evaluation on the WeatherBench South Pacific subset (2000–2004) shows that these hybrids yield measurable improvements in short-range forecast skill while maintaining physical consistency. The South Pacific region was chosen to ensure a consistent meteorological setting and to enable computationally efficient experimentation.

\section{Neural Network Backbones for Spatiotemporal Forecasting}
\label{sec:neural_backbones}

Recent advances in data-driven weather forecasting have in part relied on neural architectures adapted from computer vision, which treat atmospheric states as videos where spatial dimensions $(H,W)$ and variables (channels $C$) evolve over time. These models learn spatiotemporal correlations directly from data but often lack inherent physical constraints. In this study, we focus on two state-of-the-art computationally efficient architectures, namely, PredFormer and IAM4VP, that serve as foundations for our hybrid models, alongside the established physics-informed baseline WeatherGFT.

PredFormer adapts the Transformer architecture for video prediction by factorizing the computationally expensive self-attention mechanism. Rather than applying attention jointly over all spatiotemporal tokens, it employs two branches: one attends to the temporal evolution of each spatial location, while the other captures spatial interactions within each time step. This factorization substantially reduces computational cost while effectively representing atmospheric patterns. Its compact architecture and high accuracy make it a suitable candidate for enhancement with physical constraints.

IAM4VP \citep{imvp} addresses the trade-off between autoregressive models (prone to error accumulation) and non-autoregressive models (which may lack temporal coherence). It employs a semi-autoregressive strategy, encoding a full sequence of observed frames but decoding forecasts one step at a time. A key innovation is its dual-queue mechanism, which blends ground-truth inputs with autoregressive predictions, thereby anchoring the model to real data at longer horizons. Built with modern ConvNeXt blocks, IAM4VP achieves strong accuracy with only 12M parameters, making it highly computationally efficient.

A Physics-Informed Baseline WeatherGFT \citep{WEATHER_GFT} represents a pioneering hybrid approach, interleaving Transformer blocks with a PDE solver that integrates governing atmospheric equations over 15-minute intervals. This design enforces physical consistency by construction but originally employed a chain of 24 specialized modules, resulting in uneven gradient updates across lead times and overfitting at specific horizons. In addition, its solver was restricted by a stringent Courant–Friedrichs–Lewy (CFL) condition, reducing computational efficiency.

Building on these foundations, we combine the efficient spatiotemporal learning of PredFormer and IAM4VP with the physical consistency principle of WeatherGFT. The next sections describe enhancements to the WeatherGFT framework and the integration of its improved physical core into these neural backbones.

\section{Training Data and Experimental Setup}
\label{sec:trainingdata}
In this section we describe a procedure of neural network training.
All experiments are conducted using the WeatherBench dataset \citep{Rasp2020}, which provides reanalysis-derived atmospheric fields on a regular grid. 
To focus on a meteorologically coherent domain, the South Pacific subset is extracted, covering the region from 5.625°N–39.375°S and 174.375°E–95.625°W. 
The dataset is restricted to the years 2000–2004. Data from 2000–2003 are used for training, and the year 2004 is reserved for independent evaluation.  

The variables considered are geopotential height, temperature, the zonal (\(u\)) and meridional (\(v\)) wind components, and specific humidity at levels 50, 100, 150, 200, 250, 300, 400, 500, 600, 700, 850, 925, 1000~hPa, following the standard WeatherBench benchmarks. 
All variables are taken with the angular resolution of 1.40525° making a \(128 \times 256\) latitude–longitude grid and sampled at 60 min temporal resolution.  

More details on model parameters can be found in Appendix~\ref{sec:architecture_details_appendics}

\section{Effects of Physical Constraints on Neural Weather Prediction Architectures}
\label{sec:architecture}
This section presents a systematic analysis of how the architecture and placement of hybrid modules, where physical constraints are integrated with neural outputs, affect the performance of neural weather forecasting models.
Our aim is twofold: (i) to assess which configurations of the hybrid unit yield consistent improvements under different design choices, and (ii) to provide guidelines for avoiding overfitting and error accumulation when integrating physical priors into data-driven models.

\subsection{Transition to a Single Hybrid Block Architecture}

\begin{figure}[h]
\begin{center}
\includegraphics[width=1.0\textwidth]{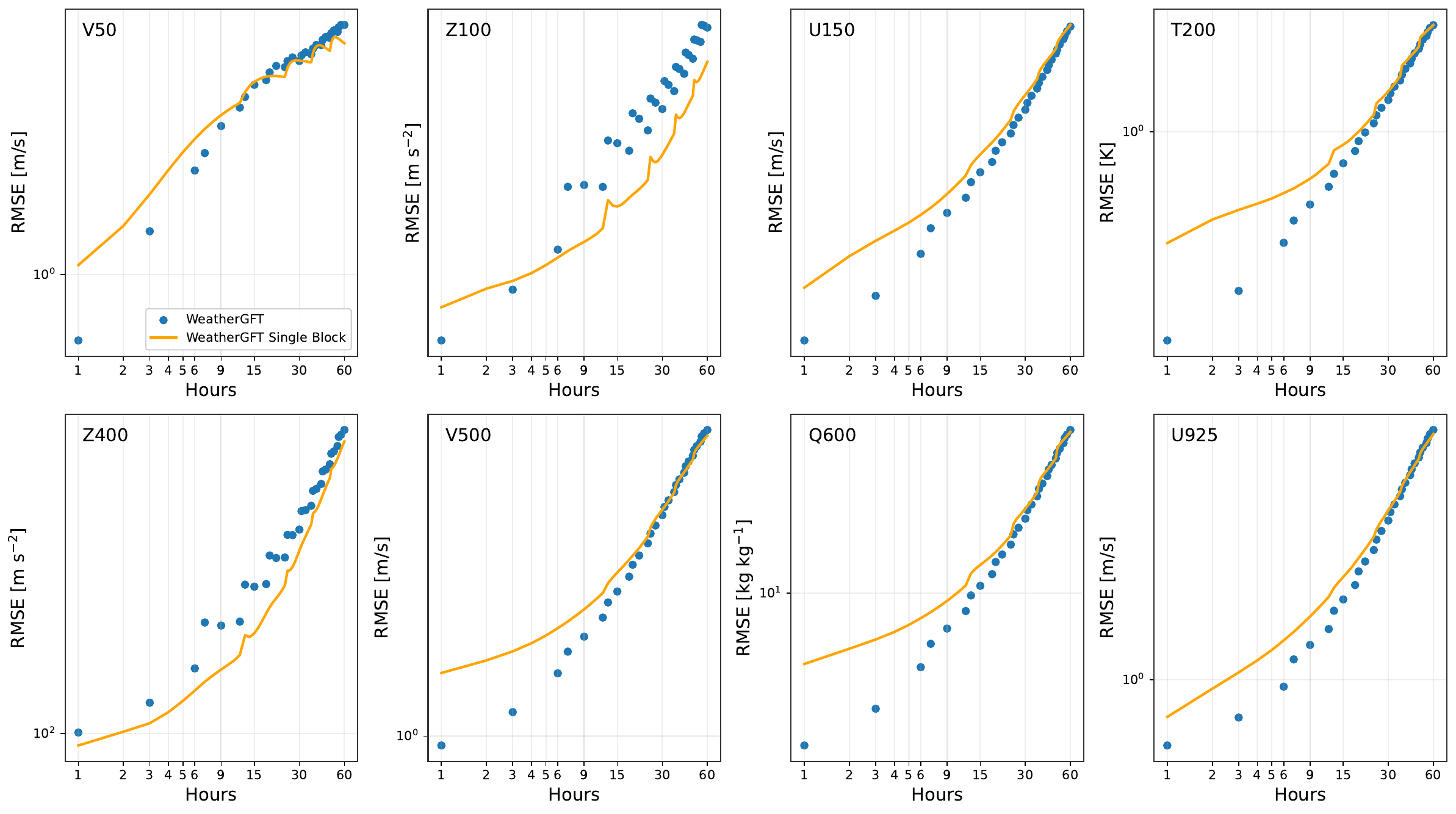} 
\end{center}
\caption{Comparison between the original \textsc{WeatherGFT} and the improved generalization of the single-block variant. Single-block variation achieves similar or better quality on long-term predictions despite having 10 times less parameters. WeatherGFT results are plotted only at the trained horizons (1, 3, and 6 h of each forecast cycle, i.e. 1, 3, 6, 7, 9, 12 h, etc.), since predictions at intermediate steps (e.g. 2, 4, 5 h) degrade severely due to overfitting and fall outside the plotted range.}
\label{fig:overfitting_gft}
\end{figure}

The original \textsc{WeatherGFT} employs 24 sequential hybrid blocks, each responsible for a 15-min forecast interval. 
Because training is supervised only at fixed horizons (1, 3, and 6~h), blocks operating at longer lead times receive fewer gradient updates than those at shorter ranges. 
This imbalance produces overfitting near the trained horizons and degraded performance at intermediate steps (e.g.\ 2–5~h), as can be seen on Figure \ref{fig:overfitting_gft}.

To address this, a simplified configuration was developed in which a \emph{single} hybrid block is applied autoregressively to generate forecasts of arbitrary length. Details on the difference from the original architecture are presented in the end of Appendix \ref{sec:appenxid_original_hybrid_block}.
This ensures that every forecast step contributes equally to parameter updates and removes discontinuities caused by disjoint training coverage across blocks. 
The consolidated block contains six PDE kernels and two neural modules, arranged to simulate a 60-min forecast via two successive 30-min steps. 
Its parameter count is approximately twice that of a single 15-min block, providing sufficient capacity while remaining an order of magnitude smaller than the full 24-block system.

The single-block design eliminates horizon-specific overfitting and substantially reduces parameter count, while still delivering accuracy comparable to the full 24-block chain at longer lead times. At the same time, the original WeatherGFT retains an advantage in some early forecast hours, where its larger capacity yields slightly better short-range skill. The new approach offers greater efficiency and more consistent accuracy across horizons, though at the expense of reduced performance in the short range. These trade-offs motivate the exploration of architectures that can exploit multiple temporal inputs, as discussed in the following subsection.

\subsection{Numerical Integration Enhancements}

Forecast errors in the original \textsc{WeatherGFT} integration arise primarily from the discretization scheme rather than deficiencies in the underlying physical equations. 
Although the five-point centred stencil is formally fourth-order accurate, it behaves poorly on non-smooth fields, producing dispersive oscillations near grid boundaries. 
These appear as artificial ripples and distorted moisture or wind fronts, which accumulate over time and degrade forecast quality.

To address these shortcomings and improve stability, we introduced several targeted kernel upgrades that together allow a fourfold increase in the integration time step without compromising accuracy:

\textbf{WENO-5 for horizontal derivatives.} 
Central differences are replaced with fifth-order \emph{Weighted Essentially Non-Oscillatory} (WENO) derivatives for each prognostic variable 
$\psi \in \{u, v, T, q, Z\}$. 
Cell-face values are reconstructed using nonlinear combinations of three parabolas, with weights adapting to local smoothness to suppress oscillations near discontinuities. 
This reduces spurious wave amplitudes by a factor of three and narrows transition zones from five to three grid points.  

\textbf{Latitude-dependent ($\beta$-plane) Coriolis term.} 
The constant Coriolis parameter $f$ is replaced by a latitude-dependent expression
\[
  f(\varphi) = f_0 + \beta R \varphi, \quad \varphi \in \left[-\tfrac{\pi}{2}, \tfrac{\pi}{2}\right],
\]
where $R$ is Earth’s radius. 
This restores meridional variability in Coriolis forcing and reduces artificial ringing near the equator.  

\textbf{Subgrid-scale eddy viscosity.} 
A diffusion term $\nu_e \nabla^2 \mathbf{u}$ is added to damp short-wavelength noise. 
The Laplacian is computed using a second WENO pass. 
With $\nu_e \approx 10^4$~m$^2$\,s$^{-1}$, spectral energy above the Nyquist half-wavenumber is reduced by about 10~dB per model hour, eliminating 2–3 gridpoint artefacts and improving stability at long horizons.  

\textbf{Increased time step.} 
The original model was limited to $\Delta t=300$~s by CFL constraints. 
With WENO-5 and eddy damping, stability analysis indicated $\Delta t_{\max} \approx 1400$~s for the $32\times64$ grid. 
A step of $\Delta t=1200$~s was adopted, reducing GPU time by more than 70\% due to fewer kernel calls.  

Upgrading the discretization scheme with WENO-5 derivatives, a latitude-sensitive Coriolis term, subgrid viscosity, and an increased time step stabilises the integration and improves forecast accuracy across all prognostic variables.

\begin{figure}[h]
\begin{center}
\includegraphics[width=\linewidth]{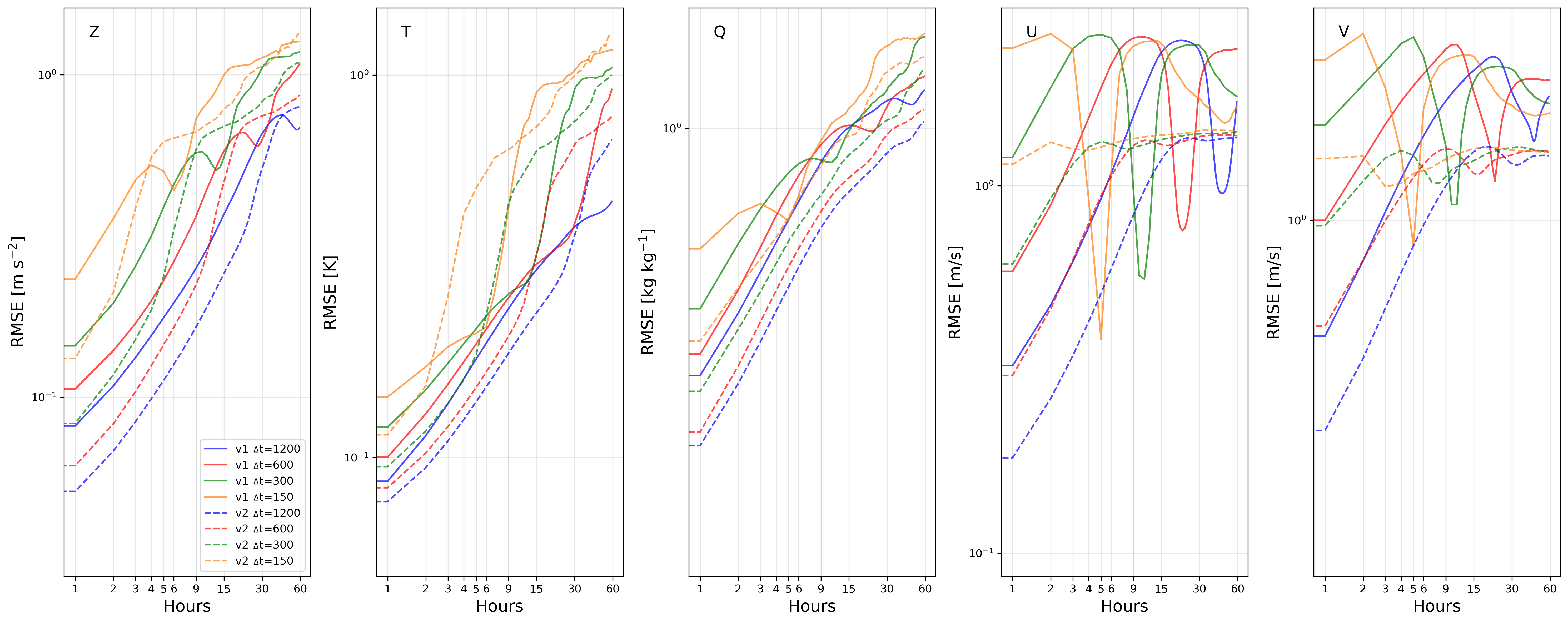}
\end{center}
\caption{Forecast error growth as a function of lead time for geopotential ($Z$), temperature ($T$), specific humidity ($Q$), zonal wind ($U$), and total wind speed ($V$). 
  Solid lines denote the original integrator; dashed lines denote the upgraded scheme. 
  Across all prognostic variables, the enhanced scheme produces slower error growth, suppresses spurious oscillations, and extends the range of low-error forecasts.}
\label{fig:all_avg}
\end{figure}

Figure~\ref{fig:all_avg} shows RMSE as a function of lead time for geopotential $Z$, temperature $T$, specific humidity $Q$, zonal wind $U$, and total wind speed $V$. 
In all cases, the new integrator (dashed lines) slows error growth compared with the previous scheme (solid lines). 
For example, $Z$ RMSE remains below 0.2 up to 6~h, compared with 3~h for the baseline, while temperature retains low errors out to nearly 12~h. 
Humidity benefits most at short leads, with an initial RMSE reduction of almost 30\% at 1~h. 
Wind forecasts also show marked improvements, with oscillatory artefacts suppressed and mid-range errors reduced. 
Overall, the new integration scheme yields consistently more accurate and stable forecasts.  

\subsection{Multi-Input–Multi-Output Architecture: PI-PredFormer}

The transition to a single hybrid block also enables the use of more expressive neural architectures, capable of exploiting not only the current atmospheric state but also recent temporal trends. 
To investigate this potential, we integrate the physics-informed module from \textsc{WeatherGFT} into the \textsc{PredFormer} architecture. 
\textsc{PredFormer} is designed for spatiotemporal forecasting, using two distinct attention mechanisms: one branch attends along the temporal axis, capturing the evolution of atmospheric states, while the other attends across locations within each time slice, representing spatial dependencies. 
This dual-branch design allows the network to combine information on both temporal progression and spatial structure, features that are central to weather prediction.

\begin{figure}[h]
\begin{center}
\includegraphics[width=0.85\textwidth]{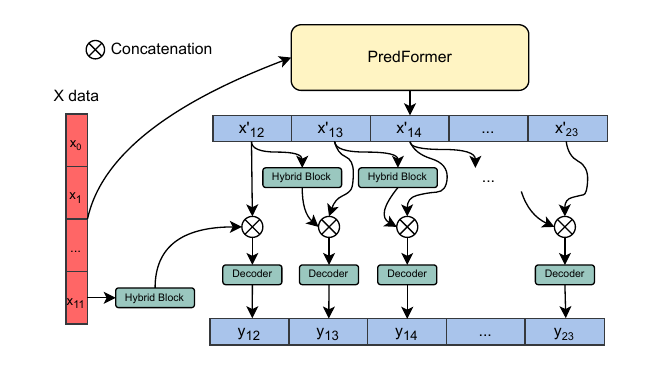} 
\end{center}
\caption{Schematic of the PI-PredFormer architecture. 
    PredFormer first processes the input sequence using temporal and spatial attention branches. 
    Physics-informed hybrid blocks then refine these forecasts, and a decoder adaptively combines neural and physical outputs to produce the final prediction.}
\label{fig:predformer_gft_scheme}
\end{figure}

The resulting hybrid, termed \textsc{PI-PredFormer}, employs a two-stage prediction process (Figure~\ref{fig:predformer_gft_scheme}). 
In the first stage, \textsc{PredFormer} generates a 12-hour forecast $X' \in \mathbb{R}^{T\times C\times H\times W}$ from a 12-hour input $X$. 
In the second stage, independent physics-informed hybrid blocks are applied to $X'$, enforcing dynamical and thermodynamical consistency. 
Finally, a decoder module merges the neural and physics-informed outputs to produce the forecast $Y$. 

Because the hybrid blocks refine forecasts at each step without relying on their own past outputs, this design avoids the autoregressive error accumulation that typically affects iterative models. 
At the same time, physical consistency is explicitly reinforced by the PDE solver. 
To accommodate 1-hour forecast intervals, the hybrid block was redesigned to consist of three 20-minute kernels, which proved more effective than a single long step (see Figure~\ref{fig:all_avg}).  

Despite these advantages, one limitation of \textsc{PI-PredFormer} is that the hybrid blocks are applied independently at each forecast step. 
This means that their influence diminishes at longer horizons: once the neural predictions begin to drift from physically plausible states, post-hoc corrections cannot fully restore consistency. 
As a result, the forecasts tend to be sharper but can occasionally appear less realistic.

Figure~\ref{fig:rmse_all_comparison} includes comparison of the performance of \textsc{PredFormer} and \textsc{PI-PredFormer}. 
The hybrid achieves clear improvements at the 1-hour lead, demonstrating the benefit of applying physical corrections to forecasts that are still anchored to observations. 
At longer horizons, however, error growth accelerates, highlighting a fundamental challenge in multi-step forecasting: early inaccuracies propagate nonlinearly through both the neural and physical branches, and post-hoc corrections cannot completely compensate. 
This finding underscores the need for tighter coupling between physical and neural components in future hybrid designs.

\subsection{Semi-Autoregressive Architecture: PI-IAM4VP}

The previous sections demonstrate both the benefits and the limitations of physics-informed modules. 
Pure autoregressive models suffer from error accumulation, while post-hoc enforcement of physical constraints cannot retroactively correct earlier inaccuracies. 
This reflects the classical trade-off between Multiple-Input–Multiple-Output (MIMO) strategies, which predict entire sequences at once, and Single-Input–Single-Output (SISO) strategies, which extend forecasts step by step.

\begin{figure}[h]
\begin{center}
\includegraphics[width=\textwidth]{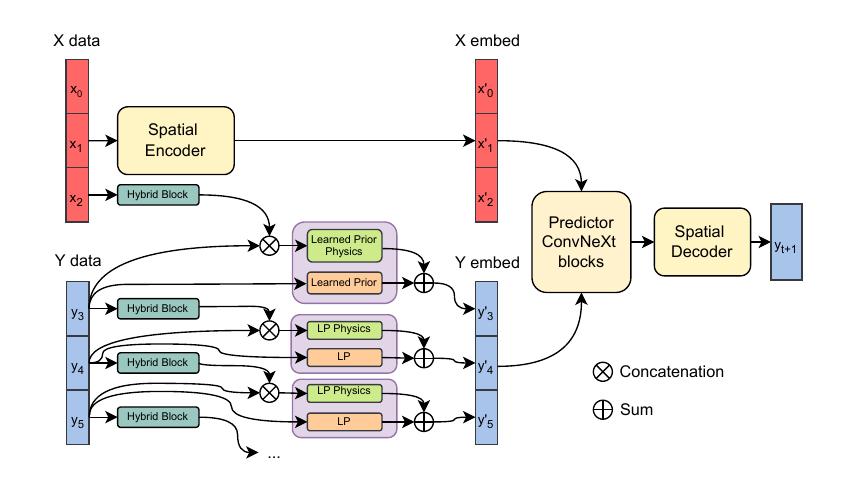}  
\end{center}
\caption{Schematic of the PI-IAM4VP architecture. 
    The input is split into a historical stream ($X$) and an autoregressive stream ($Y$). 
    Both streams are processed by hybrid blocks and combined in the decoder to produce the target forecast.}
\label{fig:PIMVP_scheme}
\end{figure}

To address these challenges, we adapt the \textsc{IAM4VP} architecture, originally developed for video prediction, to the weather forecasting problem. 
The model separates the input into two streams: 
historical observations $X \in \mathbb{R}^{T\times C\times H\times W}$ and autoregressive predictions $Y \in \mathbb{R}^{t\times C\times H\times W}$ for the interval between the last observation and the target forecast time. 
For example, to generate a 5-hour forecast, the model ingests 12 hours of reanalysis data and then uses its own predictions for the subsequent 4 hours before producing the 5-hour output. 
Intermediate steps are generated in separate forward passes. 
This design balances the stability of direct prediction from historical states with the flexibility of autoregressive extensions.

The resulting hybrid, \textsc{PI-IAM4VP}, is shown in Figure~\ref{fig:PIMVP_scheme}. Its main components are:

\begin{itemize}
    \item \textbf{Spatial Encoder.} A stack of convolutional layers interleaved with axis shifts extracts latent features from $X$. Four levels of representation are obtained, resulting in $X' \in \mathbb{R}^{4\times T\times C'\times H\times W}$.
    
    \item \textbf{Hybrid Block.} Three 20-min PDE kernels from \textsc{WeatherGFT} (without neural components) are applied, together simulating a 1-hour forecast step. 
    To accommodate the patch-based nature of the kernels, inputs are divided into $4\times 4$ subgrids. 
    
    \item \textbf{Learned Prior and LP Physics.} These modules refine neural and physics-informed features before fusion. 
    They mirror the encoder structure but use separate parameters and are shared across time steps. 
    Following the convention in \textsc{WeatherGFT}, the physical component is weighted by 0.1 in the final combination. 
    
    \item \textbf{Predictor.} A sequence of ConvNeXt~\citep{convnextv2} blocks generates next-frame predictions from the refined features. 
    
    \item \textbf{Decoder.} This combines outputs from the predictor with latent features from both $X$ and $Y$. 
    The decoding proceeds in four stages, each aligned with one of the encoder’s latent levels.
\end{itemize}

Evaluation results in Table~\ref{fig:quality_all_mid} demonstrate that embedding physics constraints within this semi-autoregressive framework yields consistent improvements across variables. 
The hybrid block enhances accuracy while maintaining stability at longer horizons. 
An interesting feature is the sharp reduction in error between the 1-hour and 2-hour forecasts. 
This may reflect the hybrid block’s convolutional operations being better tuned to the smoother synthetic fields generated by the neural module than to the noisier real initial states. 
Although duplicating the hybrid block for each forecast hour could potentially mitigate this effect, it would substantially increase model complexity. 
In this work we therefore retain a single shared block, striking a balance between accuracy and computational efficiency. 

\section{Model Evaluation and Comparison}
\label{sec:results}

Having introduced the model components, we now evaluate their forecast performance. Detailed parameters of architectures presented in this study are described in Appendix \ref{sec:architecture_details_appendics}.

Forecast quality is assessed at prediction horizons of 1, 3, 6, 12, 24, and 60 hours. Results for short-term range are presented in Table~\ref{fig:quality_all_short} and for mid-term range in  Table~\ref{fig:quality_all_mid}.

\begin{table}[t]

\caption{Root Mean Squared Error (RMSE) of forecasts at 1-hour and 3-hour horizons for all evaluated models. Non-bold entries indicate percentage increase in RMSE relative to the best model in the same column.}
\label{fig:quality_all_short}

\begin{center}
\footnotesize
\renewcommand{\arraystretch}{1.25}
\setlength{\tabcolsep}{3pt}

\begin{tabularx}{\linewidth}{@{} p{2.7cm} *{10}{>{\centering\arraybackslash}p{0.78cm}} @{}}
\multicolumn{1}{c}{\bf Model} &
\multicolumn{2}{c}{\bf z150 (m s$^{-2}$)} &
\multicolumn{2}{c}{\bf t150 (K)} &
\multicolumn{2}{c}{\bf u500 (m/s)} &
\multicolumn{2}{c}{\bf q1000 (kg kg$^{-1}$)} &
\multicolumn{2}{c}{\bf Avg.\ scaled}
\\ \hline \\

& \bf 1 h & \bf 3 h & \bf 1 h & \bf 3 h & \bf 1 h & \bf 3 h & \bf 1 h & \bf 3 h & \bf 1 h & \bf 3 h
\\ \hline \\

WeatherGFT
& +123\% & +175\% & \textbf{0.55} & +12\% & +24\% & +17\% & +33\% & +13\% & +20\% & +17\% \\

WeatherGFT Single Block
& +125\% & +168\% & +38\% & +52\% & +71\% & +66\% & +70\% & +44\% & +60\% & +50\% \\

PredFormer
& +26\% & +43\% & +20\% & +21\% & +68\% & +43\% & +89\% & +46\% & +60\% & +42\% \\

\textbf{PI-PredFormer}
& \textbf{81.9} & +63\% & \textbf{0.55} & +25\% & +53\% & +45\% & +85\% & +48\% & +50\% & +42\% \\

IAM4VP
& +47\% & +66\% & +2\% & +5\% & \textbf{1.18} & +3\% & \textbf{2.30} & \textbf{3.02} & \textbf{0.10} & +8\% \\

\textbf{PI-IAM4VP}
& +40\% & \textbf{76.1} & +4\% & \textbf{0.56} & +2\% & \textbf{1.43} & \textbf{2.30} & +1\% & \textbf{0.10} & \textbf{0.12} \\

PredRNN
& +152\% & +179\% & +60\% & +64\% & +94\% & +78\% & +61\% & +44\% & +70\% & +58\% \\

SimVP
& +138\% & +155\% & +155\% & +150\% & +145\% & +103\% & +203\% & +131\% & +140\% & +108\% \\

\end{tabularx}
\end{center}

\end{table}

\begin{table}[t]

\caption{The same as in Table~\ref{fig:quality_all_short}, but for 6-hour and 12-hour horizons}
\label{fig:quality_all_mid}

\begin{center}
\footnotesize
\renewcommand{\arraystretch}{1.25}
\setlength{\tabcolsep}{3pt}

\begin{tabularx}{\linewidth}{@{} p{2.7cm} *{10}{>{\centering\arraybackslash}p{0.78cm}} @{}}
\multicolumn{1}{c}{\bf Model} &
\multicolumn{2}{c}{\bf z150 (m s$^{-2}$)} &
\multicolumn{2}{c}{\bf t150 (K)} &
\multicolumn{2}{c}{\bf u500 (m/s)} &
\multicolumn{2}{c}{\bf q1000 (kg kg$^{-1}$)} &
\multicolumn{2}{c}{\bf Avg.\ scaled}
\\ \hline \\

& \bf 6 h & \bf 12 h & \bf 6 h & \bf 12 h & \bf 6 h & \bf 12 h & \bf 6 h & \bf 12 h & \bf 6 h & \bf 12 h
\\ \hline \\

WeatherGFT
& +141\% & +97\% & +11\% & +7\% & +12\% & +8\% & +2\% & \textbf{4.69} & +13\% & +10\% \\

WeatherGFT Single Block
& +135\% & +79\% & +41\% & +26\% & +47\% & +29\% & +26\% & +14\% & +40\% & +20\% \\

PredFormer
& +21\% & \textbf{142.51} & +11\% & \textbf{0.86} & +21\% & +5\% & +20\% & +3\% & +20\% & +5\% \\

\textbf{PI-PredFormer}
& +38\% & +11\% & +15\% & +2\% & +23\% & +8\% & +21\% & +4\% & +27\% & +10\% \\

IAM4VP
& +42\% & +23\% & +5\% & +2\% & +3\% & +3\% & +0\% & \textbf{4.69} & +7\% & +5\% \\

\textbf{PI-IAM4VP}
& \textbf{97.17} & +3\% & \textbf{0.66} & \textbf{0.86} & \textbf{1.80} & \textbf{2.41} & \textbf{3.79} & \textbf{4.69} & \textbf{0.15} & \textbf{0.20} \\

PredRNN
& +125\% & +67\% & +52\% & +33\% & +55\% & +35\% & +31\% & +18\% & +47\% & +25\% \\

SimVP
& +101\% & +40\% & +114\% & +66\% & +64\% & +30\% & +84\% & +49\% & +67\% & +30\% \\

\end{tabularx}
\end{center}

\end{table}

For geopotential height and temperature, the best-performing models are those with embedded physics modules (PI-PredFormer, WeatherGFT, and PI-IAM4VP), underscoring the value of PDE constraints at very short leads. 
For wind and humidity, however, the underlying neural architecture proves more decisive: IAM4VP achieves the strongest scores. 

At intermediate horizons, PI-IAM4VP delivers the lowest mean errors across most variables, followed by its baseline IAM4VP and by PredFormer. 
This indicates that physics-informed corrections remain effective up to about half a day, especially when coupled with a semi-autoregressive design. 

At longer horizons, purely data-driven models (notably PredFormer) take the lead. 
Here, accumulated forecast errors outweigh the corrective effect of the simplified physics module. 
Nonetheless, PI-IAM4VP consistently outperforms its baseline, showing that physical constraints still provide benefits relative to the same neural architecture without them. 
In contrast, PI-PredFormer lags behind plain PredFormer at these leads, reflecting the limitations of post-hoc corrections when neural states have already drifted.

\begin{figure}[h]
\begin{center}
\includegraphics[width=\textwidth]{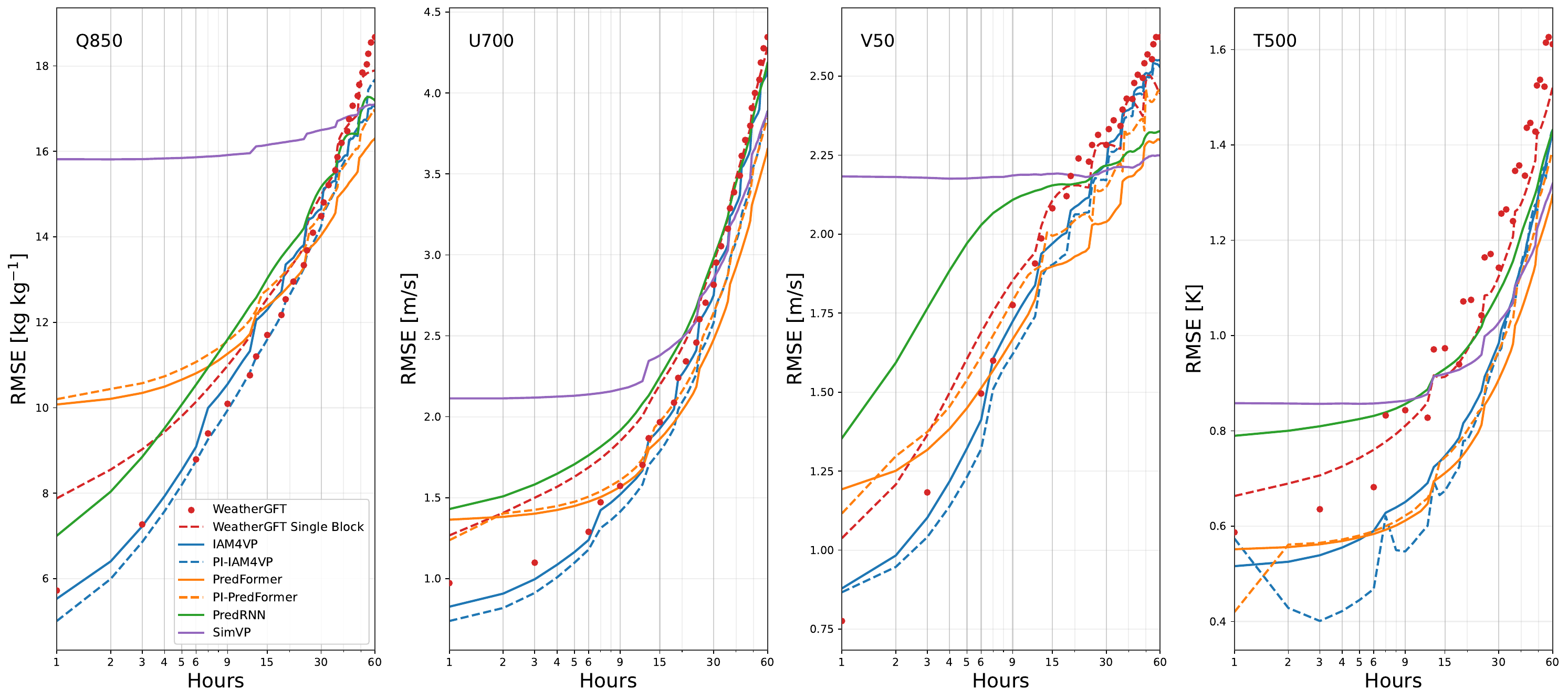}
\end{center}
\caption{Comparative evaluation of RMSE at the Q850, U700, V50, and T500 levels for various models: WeatherGFT (red dots), WeatherGFT Single Block (red dashes), IAM4VP (blue), PI-IAM4VP (blue dashed), PredFormer (orange), PI-PredFormer (orange dashed), PredRNN (green), and SimVP (purple).}
\label{fig:rmse_all_comparison}
\end{figure}

Overall, PI-IAM4VP delivers the lowest mean error across all evaluated variables up to the 12-hour horizon, demonstrating that carefully integrated physics-informed modules yield strong improvements in short- to mid-range forecasts. For longer horizons (24 and 60 hours), plain neural architectures, e.g. PredFormer, outperform their physics-augmented counterparts, likely due to error propagation overwhelming the benefits of embedded physical constraints. These performance trends are summarized in Figure \ref{fig:rmse_all_comparison}.

The evaluation highlights distinct behaviors of the two proposed hybrid architectures. 
PI-PredFormer performs best at very short lead times (1~h), where physical corrections are applied directly to observation-based states. 
This indicates that post-hoc enforcement of physical laws can stabilise forecasts when the neural predictions remain close to reality. 
However, the same mechanism becomes less effective at longer horizons: once neural trajectories diverge from physically plausible states, subsequent corrections cannot fully restore consistency. 
This leads to faster error growth beyond 6–12~h, despite initial improvements.  

PI-IAM4VP, by contrast, shows more balanced performance across variables and horizons. 
Its semi-autoregressive design allows the model to incorporate both historical observations and its own predictions, while the embedded hybrid block enforces dynamical constraints at each iteration. 
This architecture consistently improves upon its baseline up to 12~h and maintains stability at longer ranges. 
The results suggest that tighter coupling between physical modules and iterative neural forecasting is advantageous for medium-range skill.   

\section{Conclusion}
\label{sec:conclusion}

This study advances the integration of physics-informed components into neural architectures for short-range weather prediction. 
Building on the hybrid \textsc{WeatherGFT} framework, we proposed and evaluated three enhanced models that address key limitations in computational efficiency and generalisation:

\begin{itemize}
    \item \textbf{WeatherGFT-Single} condenses the original sequence of 24 specialised modules into a single autoregressive hybrid block. 
    This unified design eliminates overfitting to pre-defined lead times and reduces error discontinuities across horizons, while maintaining competitive accuracy with substantially fewer parameters.

    \item \textbf{PI-PredFormer} integrates the factorised spatiotemporal attention of PredFormer with post-hoc physics correction layers derived from \textsc{WeatherGFT}. 
    It achieves strongest performance at very short lead times (1–3 h), where physical constraints act directly on states still anchored in observations.

    \item \textbf{PI-IAM4VP} incorporates physical simulation into the semi-autoregressive framework of IAM4VP. 
    This model delivers consistent improvements across variables and lead times, demonstrating the value of embedding physics within iterative neural forecasting pipelines.
\end{itemize}

Quantitative evaluation across 1–60 h forecasts shows that physics-informed models enhance accuracy in the short-to-medium range (1–12 h), particularly for geopotential height and temperature. 

These findings demonstrate that carefully designed physics-informed neural networks improves accuracy, stability, and physical realism in the operationally critical 1–12 h window. 
Challenges remain: the PDE cores omit essential sub-grid processes; semi-autoregressive designs accumulate errors over long horizons; and post-hoc corrections cannot fully compensate for neural drift. 
Future work should explore tighter coupling of neural and physical components, including joint training of PDE solvers and neural backbones, adaptive weighting schemes responsive to forecast regimes, and incorporation of parameterised sub-grid physics. 
Extending evaluation to larger domains with additional forcings such as radiation and surface fluxes will be essential to bridge the gap towards robust, operational-scale hybrid forecasting.

\section*{Acknowledgments}
The work was supported by the grant for research centers in the field of AI provided by the Ministry of Economic Development of the Russian Federation in accordance with the agreement 000000C313925P4E0002 and the agreement with HSE University 139-15-2025-009. \\

This research is supported in part through computational resources of HPC facilities at HSE University.

\bibliography{iclr2026_conference}

@article{Rasp2020,
  title = {{WeatherBench: A Benchmark Data Set for Data‐Driven Weather Forecasting}},
  volume = {12},
  ISSN = {1942-2466},
  url = {http://dx.doi.org/10.1029/2020MS002203},
  DOI = {10.1029/2020ms002203},
  number = {11},
  journal = {Journal of Advances in Modeling Earth Systems},
  publisher = {American Geophysical Union (AGU)},
  author = {Rasp,  Stephan and Dueben,  Peter D. and Scher,  Sebastian and Weyn,  Jonathan A. and Mouatadid,  Soukayna and Thuerey,  Nils},
  year = {2020},
  month = nov 
}

@ARTICLE{Molteni1996-ph,
  title     = "The {ECMWF} Ensemble Prediction System: Methodology and
               validation",
  author    = "Molteni, F and Buizza, R and Palmer, T N and Petroliagis, T",
  journal   = "Q. J. R. Meteorol. Soc.",
  publisher = "Wiley",
  volume    =  122,
  number    =  529,
  pages     = "73--119",
  month     =  jan,
  year      =  1996,
  copyright = "http://onlinelibrary.wiley.com/termsAndConditions\#vor",
  language  = "en"
}

@article{VahidYousefnia2024,
  title = {A machine‐learning approach to thunderstorm forecasting through post‐processing of simulation data},
  volume = {150},
  ISSN = {1477-870X},
  url = {http://dx.doi.org/10.1002/qj.4777},
  DOI = {10.1002/qj.4777},
  number = {763},
  journal = {Quarterly Journal of the Royal Meteorological Society},
  publisher = {Wiley},
  author = {Vahid Yousefnia,  Kianusch and B\"{o}lle,  Tobias and Z\"{o}bisch,  Isabella and Gerz,  Thomas},
  year = {2024},
  month = jun,
  pages = {3495–3510}
}

@article{Grazzini2024,
  title = {Improving forecasts of precipitation extremes over northern and central {I}taly using machine learning},
  volume = {150},
  ISSN = {1477-870X},
  url = {http://dx.doi.org/10.1002/qj.4755},
  DOI = {10.1002/qj.4755},
  number = {762},
  journal = {Quarterly Journal of the Royal Meteorological Society},
  publisher = {Wiley},
  author = {Grazzini,  Federico and Dorrington,  Joshua and Grams,  Christian M. and Craig,  George C. and Magnusson,  Linus and Vitart,  Frederic},
  year = {2024},
  month = may,
  pages = {3167–3181}
}

@article{Zngl2023,
  title = {Adaptive tuning of uncertain parameters in a numerical weather prediction model based upon data assimilation},
  volume = {149},
  ISSN = {1477-870X},
  url = {http://dx.doi.org/10.1002/qj.4535},
  DOI = {10.1002/qj.4535},
  number = {756},
  journal = {Quarterly Journal of the Royal Meteorological Society},
  publisher = {Wiley},
  author = {Z\"{a}ngl,  G\"{u}nther},
  year = {2023},
  month = aug,
  pages = {2861–2880}
}

@article{Vitart2021,
  title = {Lagged ensembles in sub‐seasonal predictions},
  volume = {147},
  ISSN = {1477-870X},
  url = {http://dx.doi.org/10.1002/qj.4125},
  DOI = {10.1002/qj.4125},
  number = {739},
  journal = {Quarterly Journal of the Royal Meteorological Society},
  publisher = {Wiley},
  author = {Vitart,  Frédéric and Takaya,  Yuhei},
  year = {2021},
  month = jul,
  pages = {3227–3242}
}

@misc{WEATHER_GFT,
      title={Generalizing Weather Forecast to Fine-grained Temporal Scales via Physics-AI Hybrid Modeling}, 
      author={Wanghan Xu and Fenghua Ling and Wenlong Zhang and Tao Han and Hao Chen and Wanli Ouyang and Lei Bai},
      year={2025},
      eprint={2405.13796},
      archivePrefix={arXiv},
      primaryClass={cs.LG},
      url={https://arxiv.org/abs/2405.13796}, 
}

@misc{PredFormer,
      title={PredFormer: Transformers Are Effective Spatial-Temporal Predictive Learners}, 
      author={Yujin Tang and Lu Qi and Fei Xie and Xiangtai Li and Chao Ma and Ming-Hsuan Yang},
      year={2024},
      eprint={2410.04733},
      archivePrefix={arXiv},
      primaryClass={cs.CV},
      url={https://arxiv.org/abs/2410.04733}, 
}

@inproceedings{swin,
  title={Swin Transformer: Hierarchical Vision Transformer using Shifted Windows},
  author={Liu, Ze and Lin, Yutong and Cao, Yue and Hu, Han and Wei, Yixuan and Zhang, Zheng and Lin, Stephen and Guo, Baining},
  booktitle={Proceedings of the IEEE/CVF International Conference on Computer Vision (ICCV)},
  year={2021}
}

@article{Stevens2020,
  title = {Enhancement of shock-capturing methods via machine learning},
  volume = {34},
  ISSN = {1432-2250},
  url = {http://dx.doi.org/10.1007/s00162-020-00531-1},
  DOI = {10.1007/s00162-020-00531-1},
  number = {4},
  journal = {Theoretical and Computational Fluid Dynamics},
  publisher = {Springer Science and Business Media LLC},
  author = {Stevens,  Ben and Colonius,  Tim},
  year = {2020},
  month = may,
  pages = {483–496}
}

@misc{imvp,
      title={Implicit Stacked Autoregressive Model for Video Prediction}, 
      author={Minseok Seo and Hakjin Lee and Doyi Kim and Junghoon Seo},
      year={2023},
      eprint={2303.07849},
      archivePrefix={arXiv},
      primaryClass={cs.CV},
      url={https://arxiv.org/abs/2303.07849}, 
}

@misc{convnextv2,
      title={ConvNeXt V2: Co-designing and Scaling ConvNets with Masked Autoencoders}, 
      author={Sanghyun Woo and Shoubhik Debnath and Ronghang Hu and Xinlei Chen and Zhuang Liu and In So Kweon and Saining Xie},
      year={2023},
      eprint={2301.00808},
      archivePrefix={arXiv},
      primaryClass={cs.CV},
      url={https://arxiv.org/abs/2301.00808}, 
}

@inproceedings{PredRNN,
author = {Wang, Yunbo and Long, Mingsheng and Wang, Jianmin and Gao, Zhifeng and Yu, Philip S.},
title = {PredRNN: recurrent neural networks for predictive learning using spatiotemporal LSTMs},
year = {2017},
isbn = {9781510860964},
publisher = {Curran Associates Inc.},
address = {Red Hook, NY, USA},
booktitle = {Proceedings of the 31st International Conference on Neural Information Processing Systems},
pages = {879–888},
numpages = {10},
location = {Long Beach, California, USA},
series = {NIPS'17}
}

@InProceedings{simvp,
    author    = {Gao, Zhangyang and Tan, Cheng and Wu, Lirong and Li, Stan Z.},
    title     = {SimVP: Simpler Yet Better Video Prediction},
    booktitle = {Proceedings of the IEEE/CVF Conference on Computer Vision and Pattern Recognition (CVPR)},
    month     = {June},
    year      = {2022},
    pages     = {3170-3180}
}

@article{Han2022,
  title = {A short-term wind speed prediction method utilizing novel hybrid deep learning algorithms to correct numerical weather forecasting},
  volume = {312},
  ISSN = {0306-2619},
  url = {http://dx.doi.org/10.1016/j.apenergy.2022.118777},
  DOI = {10.1016/j.apenergy.2022.118777},
  journal = {Applied Energy},
  publisher = {Elsevier BV},
  author = {Han,  Yan and Mi,  Lihua and Shen,  Lian and Cai,  C.S. and Liu,  Yuchen and Li,  Kai and Xu,  Guoji},
  year = {2022},
  month = apr,
  pages = {118777}
}
\bibliographystyle{iclr2026_conference}

\appendix
\section{Appendix}

\subsection{Comparisons between new architectures and its original variants}
\label{sec:appendix_one_on_one}

\begin{figure}[h]
\begin{center}
\includegraphics[width=\textwidth]{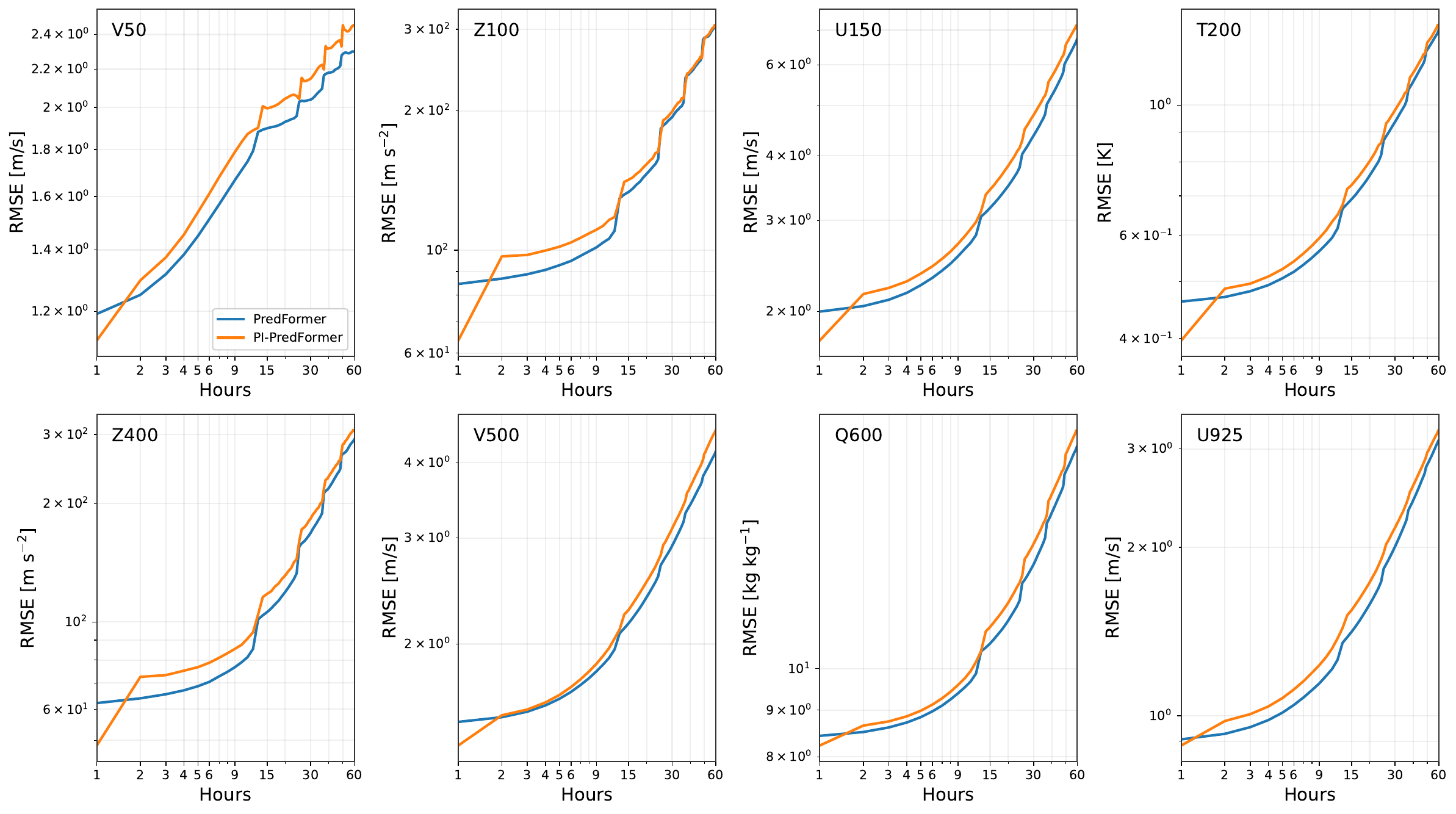}  
\end{center}
\caption{Forecast accuracy of PredFormer and PI-PredFormer in terms of RMSE for geopotential, temperature, wind, and humidity. 
    PI-PredFormer achieves improved accuracy at the 1-hour horizon, when true observations are still available as input. 
    At longer lead times, however, errors grow faster as PDE corrections are applied to increasingly inconsistent neural states.}
\label{fig:quality_predformer}
\end{figure}

\begin{figure}[h]
\begin{center}
\includegraphics[width=\textwidth]{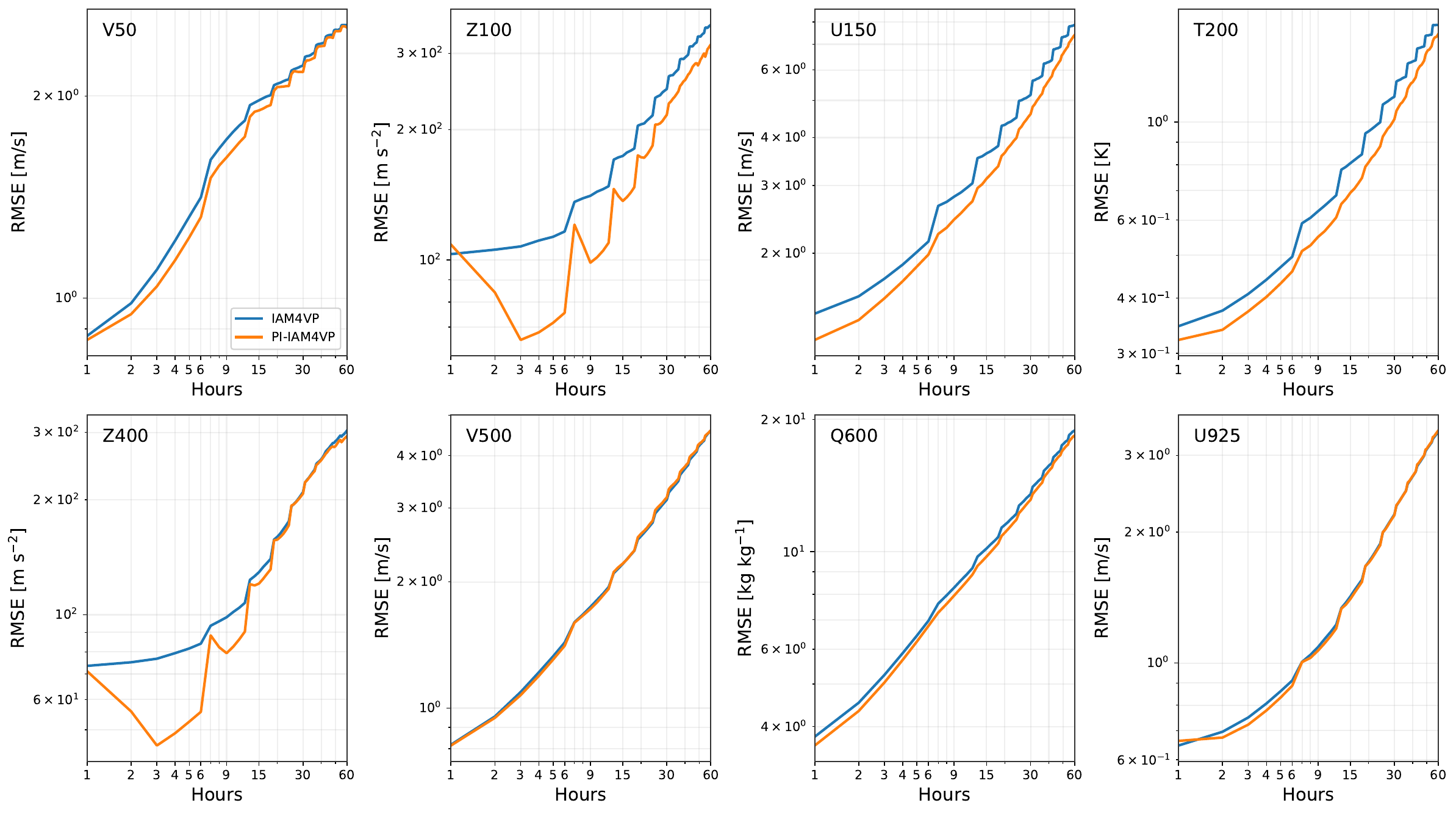} 
\end{center}
\caption{Forecast accuracy of IAM4VP and PI-IAM4VP in terms of RMSE for geopotential, temperature, wind, and humidity. 
    Across all variables and levels, the physics-informed variant provides consistently lower errors without degrading performance at longer horizons.}
\label{fig:imvp_vs_pimvp_only}
\end{figure}

\subsection{Hybrid Block in WeatherGFT}
\label{sec:appenxid_original_hybrid_block}

In the WeatherGFT architecture, each hybrid block combines outputs from the neural and physical branches before applying the PDE simulation. This summation originally uses fixed weights:
\begin{equation}
S = 0.1 \cdot P + 0.9 \cdot N, \label{0109}
\end{equation}
where $P$ and $N$ represent the outputs of the physical and neural components respectively, and $S$ is the resulting tensor passed to the PDE core.
The original implementation did not justify the choice of 0.1 and 0.9 weights in Eq. (\ref{0109}). To evaluate the impact of these coefficients, we introduced a learnable tensor $\gamma \in \mathbb{R}^{C \times H \times W}$, yielding a dynamic combination:
\begin{equation}
S = \gamma \cdot P + (\mathbf{1} - \gamma) \cdot N,
\end{equation}
where the multiplication is element-wise. A separate $\gamma$ is used for each hybrid block.

\begin{figure}[h]
\begin{center}
\includegraphics[width=1.0\textwidth]{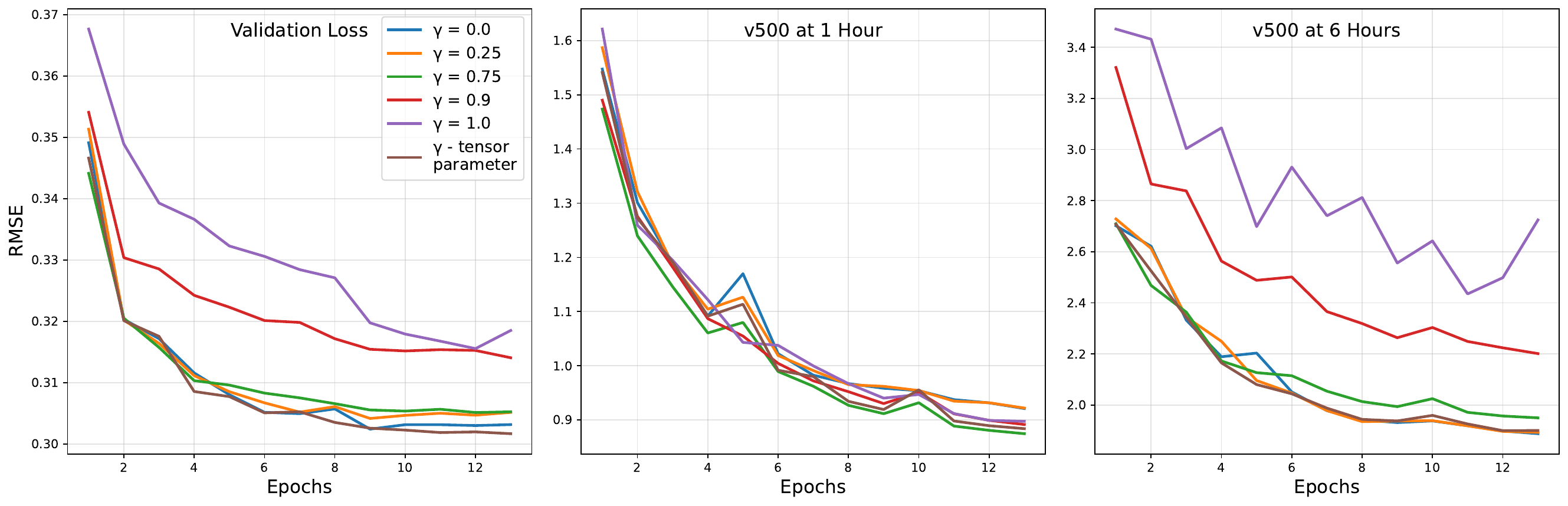} 
\end{center}
\caption{Effect of varying the weight of the physical outputs, $\gamma$, on forecast errors: fixed $\gamma$ values and the dynamic learnable $\gamma$ tensor across epochs.}
\label{fig:diff_coefs_plots}
\end{figure}

Models are trained using the global WeatherBench dataset (2000--2003) and validated with 2004 data. Results showed that learning $\gamma$ did not improve prediction accuracy. This outcome is attributed to the neural convolution already having sufficient capacity to adapt the scaling internally. However, increasing the fixed weight of the physical branch (i.e., setting $\gamma$ closer to 1) led to performance degradation, underscoring the critical role of neural corrections (see Figure~\ref{fig:diff_coefs_plots}).

Additional analysis of the trained $\gamma$ values, averaged over spatial dimensions and grouped by variable channel, revealed coefficients clustering around 0.5 - indicating a roughly equal contribution from physics and neural paths across the five PDE-governed variables.

\begin{figure}[h]
\begin{center}
\includegraphics[width=0.5\textwidth]{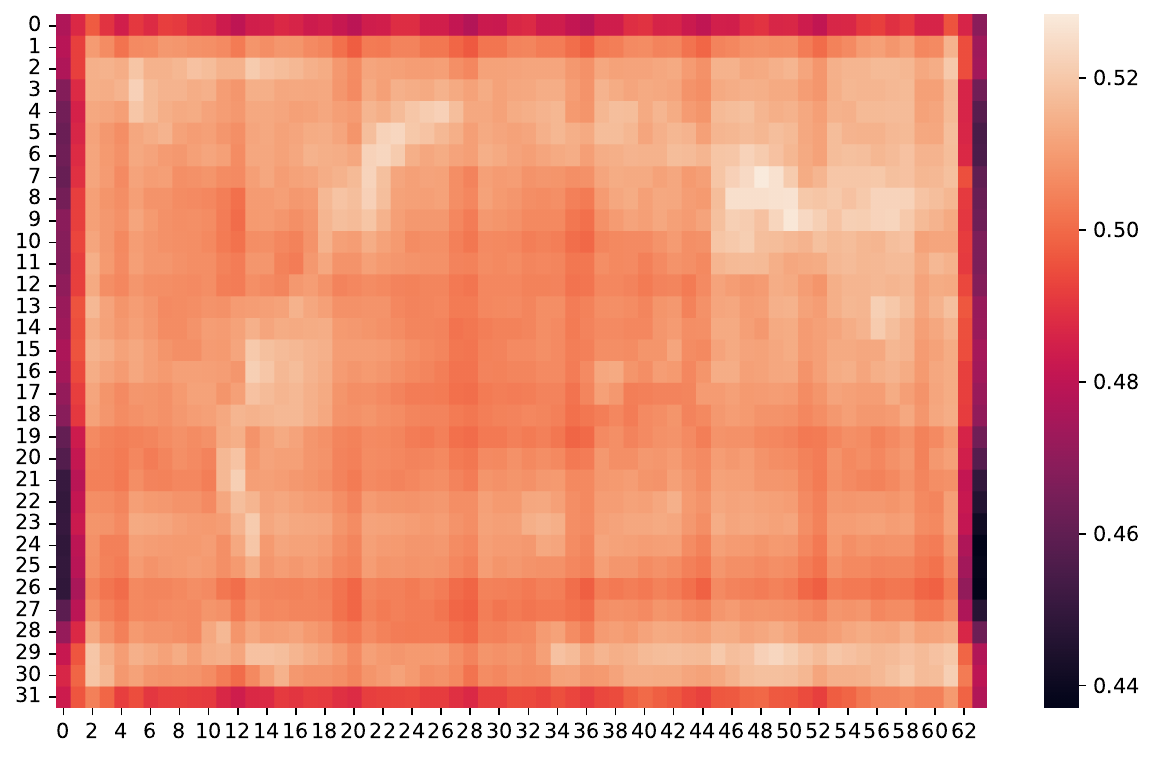}
\end{center}
\caption{Heatmap of $\gamma$ averaged over channels $C$. Larger values indicate greater contribution from the physical branch.}
\label{fig:physics_imp_plot}
\end{figure}

The spatial averaging of the values of $\gamma$ in the channels highlighted the reduced physical influence near the boundaries of the map (see Figure \ref{fig:physics_imp_plot} ). This effect stems from the limitations of discretized PDE solvers, which lack full neighborhood information at the edges. For example, the northernmost latitude cannot access data beyond the grid, which would be interpolated in a continuous PDE framework. The discrepancy becomes more pronounced in later hybrid blocks, where prediction errors accumulate and neural corrections dominate.

\begin{figure}[h]
\begin{center}
\includegraphics[width=1.0\textwidth]{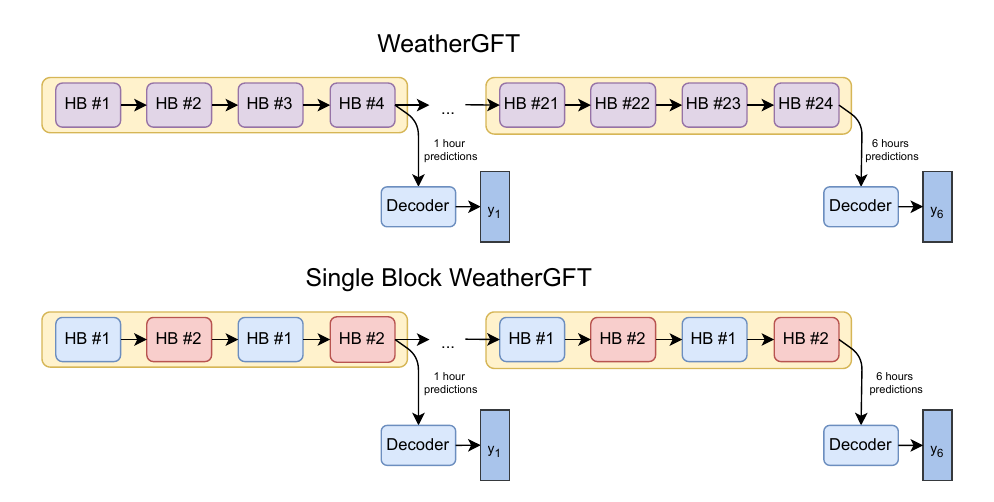} 
\end{center}
\caption{Schematic of the original \textsc{WeatherGFT} and the single-block variant. New architecture uses 2 Hybrid Blocks (HB) instead of 24, significantly reducing number of model parameters. Single consolidated Hybrid Block is implemented as two original Hybrid Blocks to accommodate the Shifting Windows attention mechanism, as described in \citep{swin}.}
\label{fig:HB_switch}
\end{figure}

This Hybrid Block is used as basis for novel Single Hybrid Block Architecture, as displayed on Figure \ref{fig:HB_switch}.

\subsection{Architecture details of presented models}
\label{sec:architecture_details_appendics}

All architectures considered in this study, including both the baseline neural networks and our physics-informed extensions are in Table~\ref{tab:all_models}.

Two established data-driven models were selected as baselines. SimVP \citep{simvp} is a lightweight convolutional network for next-frame spatiotemporal prediction, while PredRNN \citep{predrnn} is a recurrent architecture with spatiotemporal memory, widely used as a benchmark in neural weather forecasting.

All neural components are trained using the AdamW optimizer with cosine annealing learning-rate schedule. 
The initial learning rate is set to $10^{-4}$ for PredFormer-based models and $5 \cdot 10^{-4}$ for the other architectures. 
Batch size is 2 for PredFormer-based models and 16 for IAM4VP-based models, training is performed for 20 epochs with early stopping on plateau. 
Weight decay is not applied.  

The physics-informed modules are integrated directly into the training loop: after the neural network produces forecasts at each step, the PDE solver applies a correction before the loss is computed. 
The objective function combines mean squared error across all prognostic variables with the implicit enforcement of physical consistency.  

All experiments are conducted on NVIDIA GPUs with mixed precision to reduce memory footprint. 
Training a single model required approximately 150 GPU hours, while inference runs in real time on a single GPU.  

\begin{table}[t]

\caption{Overview of backbone and hybrid architectures evaluated in this study}
\label{tab:all_models}

\begin{center}
\footnotesize
\renewcommand{\arraystretch}{1.15}
\setlength{\tabcolsep}{4pt}

\begin{tabularx}{\linewidth}{@{} p{2.7cm} c p{2.6cm} p{2.8cm} Y @{}}
\multicolumn{1}{c}{\bf Model} &
\multicolumn{1}{c}{\bf Parameters} &
\multicolumn{1}{c}{\bf Type} &
\multicolumn{1}{c}{\bf Forecasting strategy} &
\multicolumn{1}{c}{\bf Key features}
\\ \hline \\

PredFormer \citep{PredFormer} & 108M & Transformer & MIMO &
Factorized temporal/spatial attention; efficient spatiotemporal learning \\

PI-PredFormer (this work) & 108M & Hybrid (Transformer+PDE) &
MIMO + post-hoc correction &
PredFormer with physics block applied every 15 min \\

\\
IAM4VP \citep{imvp} & 12M & ConvNeXt semi-autoregressive &
Semi-autoregressive &
Dual-queue blending of observations and predictions; compact and efficient \\

PI-IAM4VP (this work) & 12M & Hybrid (ConvNeXt+PDE) &
Semi-autoregressive + physics &
IAM4VP with embedded hourly physics correction \\

\\
WeatherGFT \citep{WEATHER_GFT} & 256M & Hybrid (Transformer+PDE) &
SISO (24-step chain) &
Original architecture: 24 specialised NN--PDE modules (15 min each) \\

WeatherGFT-Single (this work) & 99M & Hybrid (Transformer+PDE) &
SISO (autoregressive) &
Single hourly hybrid block, reused iteratively; reduced overfitting \\

\\
PredRNN \citep{predrnn} & 41M & Recurrent NN & SISO (autoregressive) &
Popular spatiotemporal RNN baseline \\

SimVP \citep{simvp} & 5M & CNN baseline & MIMO &
Lightweight fully convolutional model \\

\end{tabularx}
\end{center}

\end{table}

\subsection{Additional comparison between models}
\label{sec:additional_comparison}

\begin{figure}[h]
\begin{center}
\includegraphics[width=\textwidth]{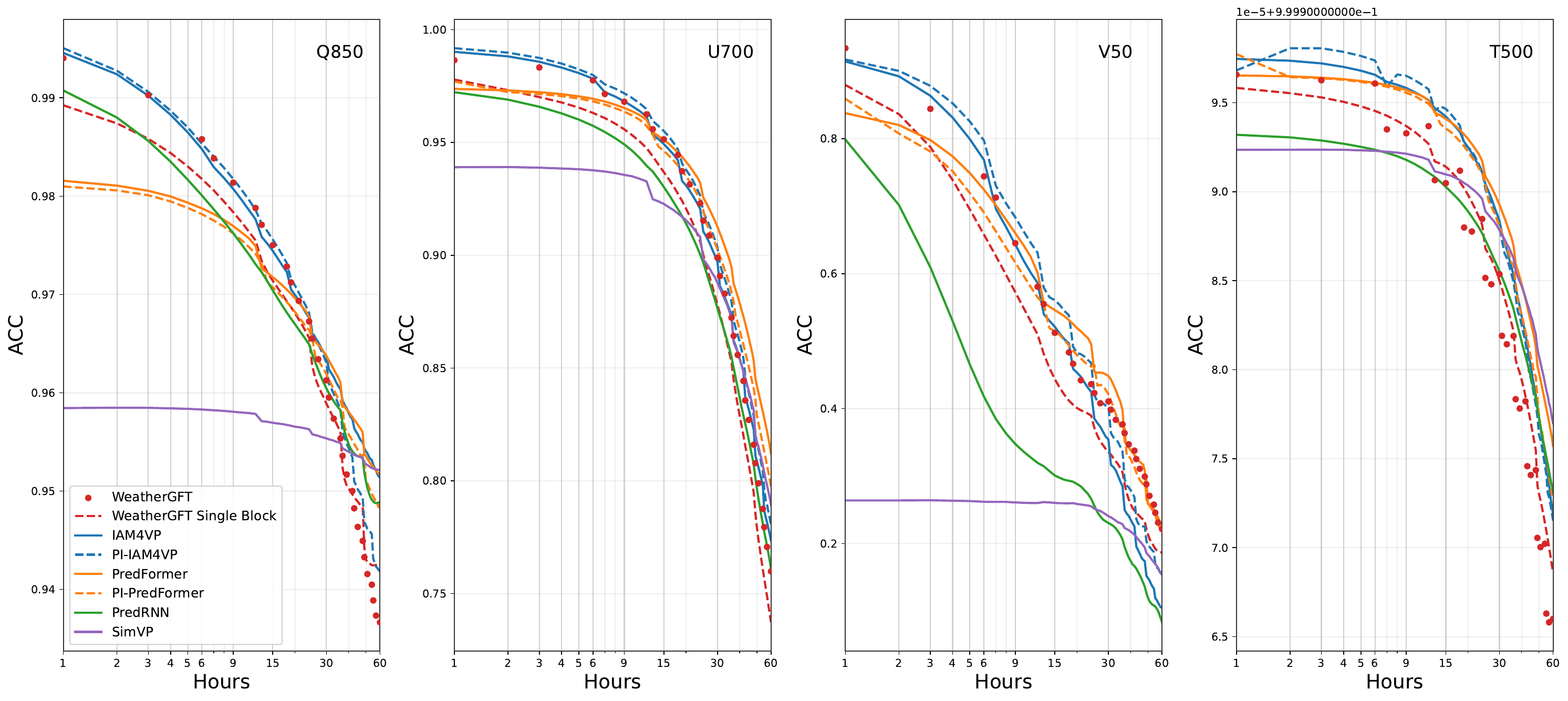} 
\end{center}
\caption{Comparative evaluation of ACC at the Q850, U700, V50, and T500 for various models: WeatherGFT (red dots), WeatherGFT Single Block (red dashes), IAM4VP (blue), PI-IAM4VP (blue dashed), PredFormer (orange), PI-PredFormer (orange dashed), PredRNN (green), and SimVP (purple).}
\label{fig:imvp_vs_pimvp}
\end{figure}

The anomaly correlation coefficient (ACC) complements RMSE by quantifying spatial pattern fidelity relative to climatology. 
Figure~\ref{fig:imvp_vs_pimvp} shows that physics-informed variants maintain higher ACC values up to 12~h, especially for wind and geopotential, indicating improved phase alignment with reanalysis. 
At longer ranges, ACC values converge across models.

\begin{figure}[h]
\begin{center}
\includegraphics[width=\textwidth]{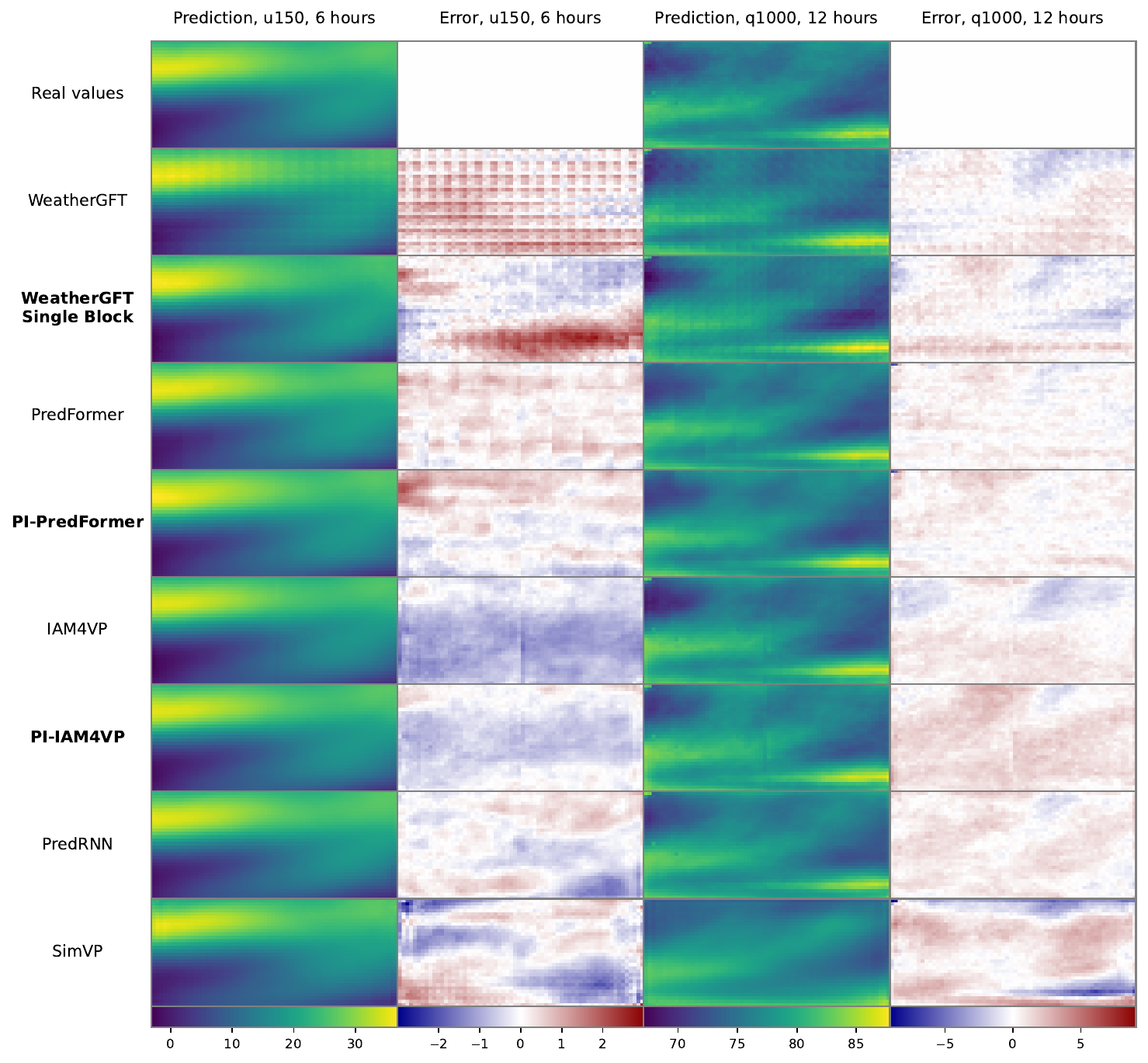}
\end{center}
\caption{Example forecast maps at 6 h and 12 h for zonal wind at 150 hPa ($u150$) and specific humidity at 1000 hPa ($q1000$). 
Purely neural baselines exhibit grid artefacts and excessive smoothing, while hybrid models (WeatherGFT-Single, PI-PredFormer, PI-IAM4VP) produce sharper gradients and more realistic synoptic patterns. 
These maps illustrate the spatial fidelity benefits of embedding physical constraints.}
\label{fig:many_maps}
\end{figure}

The map-based comparison in Figure~\ref{fig:many_maps} illustrates these findings for forecasts of zonal wind at 150~hPa and specific humidity at 1000~hPa. 
Hybrid models such as WeatherGFT and PI-IAM4VP better reproduce realistic synoptic structures and moisture gradients, whereas purely neural baselines show grid artifacts or excessive smoothing. 
These visual diagnostics confirm that hybrid approaches achieve a favorable balance between physical fidelity and adaptability to data.

At longer horizons (24–60 h), purely data-driven models—especially PredFormer—outperform their hybrid counterparts as accumulated errors outweigh the benefits of simplified physical constraints. 
Spatial analyses further reveal that hybrid models produce more realistic synoptic structures, with sharper gradients and fewer grid artefacts than purely neural baselines.

From a meteorological perspective, these findings underline the potential and the limitations of hybrid approaches. 
Physics-informed corrections are most beneficial when applied frequently and when forecasts remain anchored to observational information. 
For extended horizons, however, purely data-driven models currently remain more competitive, reflecting both the incompleteness of the simplified PDE cores and the challenges of mitigating error accumulation. This is further supported by the results presented in Table~\ref{fig:quality_all_long}, comparing the models at 24 and 60 hours prediciton range.

\begin{table}[t]

\caption{The same as in Table~\ref{fig:quality_all_short}, but for 24-hour and 60-hour horizons}
\label{fig:quality_all_long}

\begin{center}
\footnotesize
\renewcommand{\arraystretch}{1.25}
\setlength{\tabcolsep}{3pt}

\begin{tabularx}{\linewidth}{@{} p{2.7cm} *{10}{>{\centering\arraybackslash}p{0.78cm}} @{}}
\multicolumn{1}{c}{\bf Model} &
\multicolumn{2}{c}{\bf z150 (m s$^{-2}$)} &
\multicolumn{2}{c}{\bf t150 (K)} &
\multicolumn{2}{c}{\bf u500 (m/s)} &
\multicolumn{2}{c}{\bf q1000 (kg kg$^{-1}$)} &
\multicolumn{2}{c}{\bf Avg.\ scaled}
\\ \hline \\

& \bf 24 h & \bf 60 h & \bf 24 h & \bf 60 h & \bf 24 h & \bf 60 h & \bf 24 h & \bf 60 h & \bf 24 h & \bf 60 h
\\ \hline \\

WeatherGFT
& +65\% & +26\% & +14\% & +19\% & +10\% & +15\% & +5\% & +14\% & +12\% & +18\% \\

WeatherGFT Single Block
& +51\% & +17\% & +25\% & +23\% & +23\% & +19\% & +12\% & +9\% & +19\% & +18\% \\

PredFormer
& \textbf{218.93} & \textbf{457.50} & \textbf{1.09} & \textbf{1.73} & \textbf{3.34} & \textbf{5.70} & \textbf{5.51} & \textbf{7.58} & \textbf{0.26} & \textbf{0.40} \\

\textbf{PI-PredFormer}
& +7\% & +3\% & +5\% & +5\% & +6\% & +6\% & +3\% & +5\% & +4\% & +5\% \\

IAM4VP
& +29\% & +12\% & +10\% & +18\% & +7\% & +10\% & +5\% & +12\% & +8\% & +10\% \\

\textbf{PI-IAM4VP}
& +19\% & +11\% & +8\% & +10\% & +5\% & +7\% & +6\% & +13\% & +8\% & +10\% \\

PredRNN
& +45\% & +13\% & +26\% & +14\% & +25\% & +14\% & +12\% & +6\% & +19\% & +10\% \\

SimVP
& +24\% & +5\% & +40\% & +15\% & +14\% & +4\% & +32\% & +8\% & +15\% & +5\% \\

\end{tabularx}
\end{center}

\end{table}

\end{document}